\documentclass[]{article}
\usepackage{comment}
\usepackage{placeins}
\usepackage{graphicx}
\usepackage{bbding}
\usepackage[pdfborder={0 0 0}]{hyperref}
\usepackage[]{authblk}
\usepackage{setspace}
\usepackage{lineno}
\usepackage[utf8]{inputenc}
\usepackage{url}
\usepackage{placeins}
\usepackage{xspace}
\usepackage{authblk}

\providecommand{\keywords}[1]
{
  \small	
  \textbf{\textit{Keywords---}} #1
}

\title{Automatic Detection of Rice Disease in Images of Various Leaf Sizes}

\begin{document}

\author[1]{{Kantip Kiratiratanapruk}}
\author[1]{{Pitchayagan Temniranrat}}
\author[1]{\\{Wasin Sinthupinyo}}
\author[1]{{Sanparith Marukatat}}
\author[2]{\\{Sujin Patarapuwadol}}

\affil[1]{National Electronics and Computer Technology Center (NECTEC)
112 Phahonyothin Road, Khlong Nueng, \newline
Khlong Luang District, Pathumthani, Thailand 12120
kantip.kiratiratanapruk@nectec.or.th
}
\affil[2]{Department of Plant pathology, Faculty of Agriculture at Kamphaeng Saen, Kasetsart University, Nakorn Pathom, Thailand
sujin.p@ku.th
}

\maketitle
\begin{abstract}
Fast, accurate and affordable rice disease detection method is required to assist rice farmers tackling equipment and expertise shortages problems.
In this paper, we focused on the solution using computer vision technique to detect rice diseases from rice field photograph images.
Dealing with images took in real-usage situation by general farmers is quite challenging due to various environmental factors, and rice leaf object size variation is one major factor caused performance degradation.
To solve this problem, we presented a technique combining a CNN object detection with image tiling technique, based on automatically estimated width size of rice leaves in the images as a size reference for dividing the original input image.
A model to estimate leaf width was created by small size CNN such as 18 layer ResNet architecture model.
A new divided tiled sub-image set with uniformly sized object was generated and used as input for training a rice disease prediction model. 
Our technique was evaluated on 4,960 images of eight different types of rice leaf diseases, including blast, blight, brown spot, narrow brown spot, orange, red stripe, rice grassy stunt virus, and streak disease. 
The mean absolute percentage error (MAPE) for leaf width prediction task evaluated on all eight classes was 11.18\% in the experiment, indicating that the leaf width prediction model performed well. 
The mean average precision (mAP) of the prediction performance on YOLOv4 architecture was enhanced from 87.56\% to 91.14\% when trained and tested with the tiled dataset.
According to our study, the proposed image tiling technique improved rice disease detection efficiency.

\end{abstract}


\section{Introduction}

Rice is an important crop that provides food for more than half of the world's population. In Thailand, rice is an important economic crop and a major export item. Rice production is the primary income source of most of Thai farmers. However, tools to detect crop diseases and specialists to provide solutions are inadequate. Although there are portable kits for disease diagnostic tests, the procedure is still time-consuming and expensive. 
Moreover, to obtain precise diagnostic results, samples of the diseased plant are required to be brought to a laboratory for an examination.
Such problems and farmers' lack of knowledge about the diseases can cause wrong solutions, which may lead to disease widespread and damaging outbreaks. 
These problems motivated us to develop an effective and user-friendly rice disease diagnostic system. 

Due to users friendliness of disease diagnosis framework based on photograph image and the high performance of recent image processing technology based on Deep Learning techniques, we paid attention on applying Deep Learning techniques to plant diagnosis.
In this section we firstly made literature reviews about previous researches that applied Deep Learning techniques to plant diagnosis. We then introduced our previous researches that aimed on developing rice diseases detection chatbot system and it's problem found during real usage, that the object size variation in the images caused degradation of the disease detection performance. We also addressed about previous researches handled with this object size variation problems.

\subsection{Literature Review of Deep Learning for Plant Diagnosis}

Convolutional Neural Network, CNN \cite{Lecun_1998}, enables automatic decision and features extraction from input images to classify and/or localize objects in the image. 
CNN is an example of a deep learning technique \cite{Hinton_2006}, an artificial neural network technique with several multi-layer architecture.
Deep learning technique could deal with more complicate images classification and object detection problems.

Literature reviews on plant disease diagnosis applying deep learning technique in the recent years were presented in references \cite{Liu_Wang_2021,Li_Zhang_Wang_2021,Saleem_Potgieter_2019}.
In \cite{Liu_Wang_2021}, the authors reviewed both plant diseases and pests and discussed three technical frameworks in deep learning techniques: classification, detection, and segmentation. They presented the summary of pros and cons, solutions for each technique, and concluded trends with challenges for future plant disease research. In \cite{Li_Zhang_Wang_2021} the authors explained how deep learning was used to diagnose plant disease comprehensively in many aspects. 
Disease detection and classification performance in most research studies depended
on the collected image database and could not efficiently deal with the images that differed greatly from images in the database. 
There was still a gap in this field of research from the relatively limited database of plant disease images caused by many constraints, including season, duration, disease frequency, and weather \cite{Saleem_Potgieter_2019}. 
To create such large gallery takes a high-cost investment because disease surveys are labor-intensive and annotations of disease object areas in images require specific knowledge. 

Therefore, most research \cite{Hughes_Salathe_2015,Edna_Li_Sam_Liu_2019,Umit_Murat_Kemal_Emine_2021}, used free open public data such as Plant Village dataset \cite{Hughes_Salathe_2015} from the Kaggle website. 
PlantVillage provided large archives covering many types of plants and disease species (54,306 images from 14 plant species and 26 plant diseases), which were collected either from laboratory setups or in real conditions in cultivation fields. 
The authors in \cite{Edna_Li_Sam_Liu_2019} presented fine-tuning deep learning models to identify plant diseases by comparing various types of CNN architectures such as VGG16, InceptionV4, ResNet, and DenseNets. The experimental results showed that DenseNets was the best model and achieved an accuracy score of 99.75\%. In \cite{Umit_Murat_Kemal_Emine_2021}, the authors examined EfficientNet, another deep learning architecture, and set the the network's input size as different eight variable. By comparison with state-of-the-art CNN models, such as AlexNet, ResNet50, VGG16, and Inception V3, the most efficient model in identification of plant leaf diseases was determined. 
It was found that most previous research studied photograph images taken in an ideal setting that limited to plain backgrounds under certain conditions, cropped frame to a single leaf or limited the distance between the camera and the subject. 
Therefore, these papers mainly focused on the classification of disease species on photographs of single leaf, not on an image of detecting multiple leaves or multiple disease in actual field conditions. The latter case would not only classify the type of object but also identify the location of the object area in the image.

There were a few studies \cite{Zhou_Zhang_Chen_He_2019,Deng_Tao_Xing_2021,Chowdhury_Preetom_Mohammed_Sajid_2020,Jiang_Jie_Guannan_Fenghua_2017} that focused rice disease photographs in the actual environment. 
The authors in \cite{Zhou_Zhang_Chen_He_2019} developed a technique to solve the problem of environmental background interference on the objects of interest in the photographs. They used K-means clustering algorithm and Faster R-CNN algorithm fusion to tackle noise reduction and segmentation techniques. Their results achieved an average accuracy of 97.2\%, tested on three types of rice leaf diseases from a total of 3,010 images.
The authors in \cite{Deng_Tao_Xing_2021} tried to solve a problem of confusion and misjudgment of some diseases. They developed an ensemble model combined network submodels instead of using a single model in order to improve object detection performance. 
Experimental results were compared on five models namely ResNet-50, DenseNet-121, SE-ResNet-50, ResNeXt-50, and ResNeSt-50. The ensemble model integrated three best performance models, DenseNet-121, SE-ResNet-50, and ResNeSt-50.
Although the ensemble model's performance was highly accurate at 91\% tested on 33,026 images of six types of rice diseases, it required many parameters, which slowed down the identification process.
On a contrary, authors in \cite{Chowdhury_Preetom_Mohammed_Sajid_2020} studied five state-of-the-art CNN architectures to develop a memory-efficient architecture for mobile application development. They suggested a simple network architecture technique with fewer parameters using a new two-stage training method to develop light-weight CNN for rice plant disease and pest identification. 
Their experiments were able to achieve 93.3\% accuracy evaluated on small dataset of 1,426 self-prepared images of nine plant species from multiple rice disease species and planting areas.
In \cite{Jiang_Jie_Guannan_Fenghua_2017}, the authors aimed to develop a mobile phone application to perform real-time diagnosis for wheat disease identification and localization. They presented a weakly supervised deep learning framework on a new in-field wheat disease dataset, which consisted of 9,230 images with 7 different rice wheat disease classes. The images were collected from multiple sources to simulate a diverse set of photographic images closer to practical usage. They studied and conducted experiments based on four types of CNN architecture models in the VGG family and achieved the best accuracy at 97.95\%. They aimed to evolve into more challenging mixed cases of multiple diseases or multiple crops.

As previously mentioned, previous researches presented various approaches for handling challenges in many topic of interest. In \cite{Zhou_Zhang_Chen_He_2019, Deng_Tao_Xing_2021}, they concentrated on disease detection models in a problem of variety of disease morphology and complex background variation, while \cite{Chowdhury_Preetom_Mohammed_Sajid_2020, Jiang_Jie_Guannan_Fenghua_2017} focused on developing models capable of processing fast enough for real-time detection.
However, there were several gaps in this research topic that need to be investigated further in order to be practical.
The problem of different object sizes in images is another issue that had not been addressed in this field of research.
In this paper, we focused on the problem of objects of various sizes in images. This was one of the inevitable obstacles caused by a variety of users environment found in our previously operating rice disease diagnosis system which will be explained in detail in the next subsection. 

\subsection{Previously Proposed Rice Disease Detection Study and Object Relative Size Problems}

In our previous studies \cite{Kantip_2020,Pitchayagan_2021}, we investigated rice plant diseases in an actual rice field environment. In \cite{Kantip_2020}, we studied 6 varieties of major rice diseases found in Thailand, including blast, bacterial leaf blight, brown spot, narrow brown spot, bacterial leaf streak, and rice ragged stunt virus disease. We compared the efficiency of diseases detection by 4 different CNN architectures for object detection techniques, including Faster R-CNN \cite{Ren_He_FasterRCNN_2015}, RetinaNet \cite{Lin_Goyal_RetinaNet_2017}, YOLOv3 \cite{Redmon_Farhadi_YOLOV3_2018}, and Mask R-CNN \cite{He_Gkioxari_MaskRCNN_2017}. Their performance in mean average precision (mAP) were 70.96\%, 36.11\%, 79.19\%, and 75.92\%, respectively. We used YOLOv3 and its family as the base architecture because it showed the best performance. 

In \cite{Pitchayagan_2021}, we developed the object detection model training and refinement process to improve the detection efficiency of the previous research \cite{Kantip_2020}. 
We improved the model detection performance. Our model offered primary disease analysis images taken from actual rice fields and sent to a farmer group. Pathology experts provided detailed knowledge about the disease through the group conversation. 

\begin{figure}[hbtp!]
\begin{center}
\includegraphics[width=0.95\textwidth]{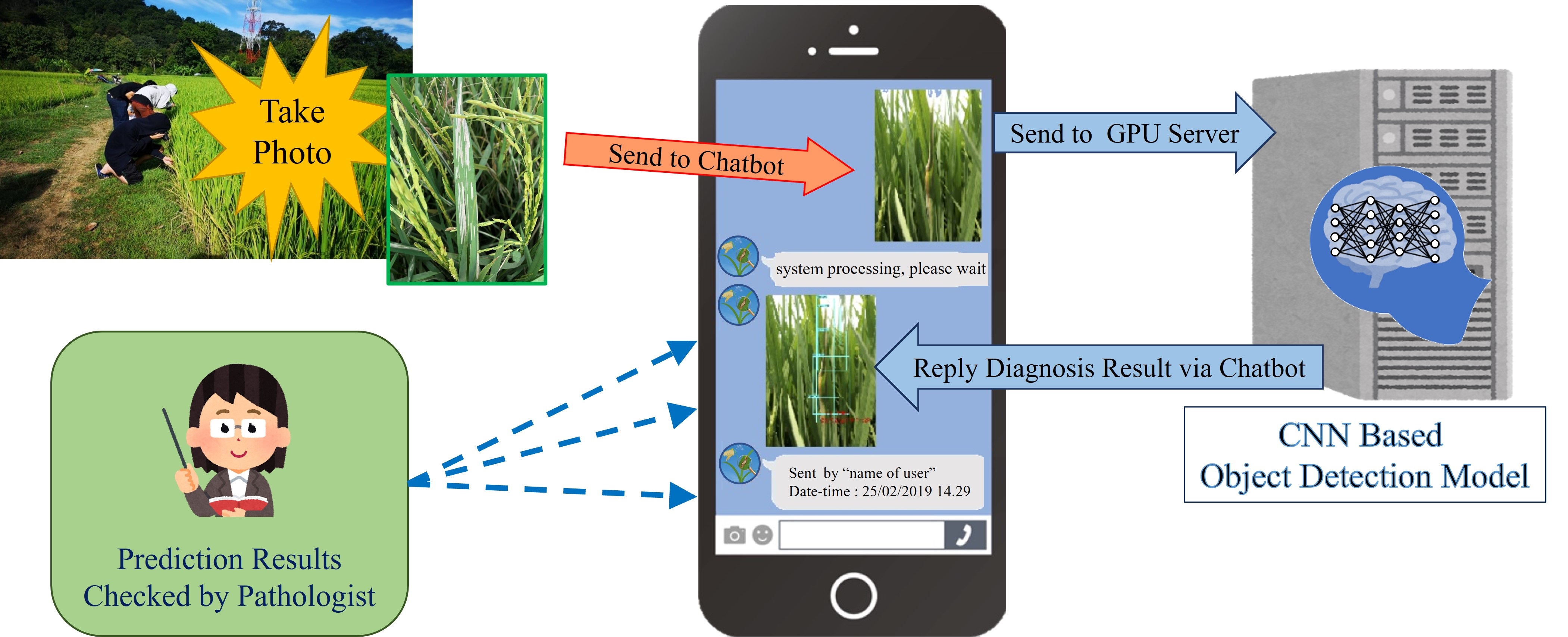}
\caption{Flow process of our rice disease Line Bot system}
\label{fig:RiceDiseaseLineBotSystem.jpg}
\end{center}
\end{figure}

A chatbot, an automatic conversation robot, was deployed to instantly diagnose  primary diseases based on images taken from paddy fields submitted in LINE application group conversations. 
We used LINE application as a communication platform due to its popularity in Thailand.
A Computer or a smart mobile phone, a camera, and internet connection were required to use the system. A workflow process of our rice disease diagnostic system is shown in Figure\ref{fig:RiceDiseaseLineBotSystem.jpg}. The images received by chatbot were sent to a server computer to diagnose rice diseases using deep learning (CNN) based object detection technique.
The diagnosis would be sent back to the LINE group by the chatbot. The LINE groups also worked as communication tools for farmers and plant pathologists, the specialists on rice disease analysis.
The specialists in the group could help to verify whether the chatbot's results are accurate and advice farmers on alternative solutions.

The disease diagnosis results from the previous model \cite{Kantip_2020} 
and knowledge obtained via group conversation log were used in further refinement process.
After the refinement process, disease prediction performance improved from 91.1\% to 95.6\%, evaluated on test dataset by Average True Positive Point. The performance on field usage was 78.86\%.
This number was acceptable considering the system had the specialists' verification.
In March 2020, we had the average of 149/31=4.8 automatic diagnosis transactions per day, 
which the specialists could check the system's diagnosis results thoroughly.
However, considering increasing users in the future and the lack of specialist, higher accuracy is required for the automatic diagnosis results.

We gathered lesson learned from the system diagnosis results and the specialists' opinions. We realized that the model had unsatisfactory results under some specific conditions. 
It was obvious to us that the rice disease object in picture data was quite different in size. This could be caused by many factors such as the size of the wound, the type of disease, the plant's growth stage, and user's distance from the object.
In the section \ref{section:expII} we made an experiment to confirm that object size variation affected degradation of the disease detection performance.
It was difficult for the general users to take pictures which had specified relative object size in the image. On the other hand, it was almost impossible and unpractical to prepare training data with sufficient object size variety that can be found in the nature.

In computer vision object detection research field, object size variety, multiple objects or extremely small object size in the image were challenging problem. 
There were several object detection techniques related to this problem \cite{Daye_M_A_2021, Xuewei_Jun_2021, Sun_Yang_He_Wu_2020}, such as pre-post processing techniques (data augmentation, network input size, image tiling), 
pre-trained model, and network architecture solutions 
(multi-scale feature map 
, multi-scale region proposal network (RPN), feature pyramid network (FPN), anchor boxes and loss function). 
These techniques interested in various directions or purposes, but they all aimed at improvement of object detection accuracy. 
However, in these researches any additional process for performance enhancement inevitably complicated the process in some way, such as memory consumption, inference time, or computation cost.

In this paper, we chose image tiling technique to solve the problem. 
Image tiling is dividing image into multiple same padding size sub-images, either overlap or non-overlap.
The divided sub-images were used as the new input images.
Unlike resizing an image, that might lose details and object information, 
the tiling technique maintained image aspect-ratio and reduced the loss of image information. We employed this strategy to improve the quality of the data used to train the detection model rather than directly modifying the structure of the model network. It is flexible in terms of model-free configuration, allowing model architecture to be replaced if a new model with better performance becomes available. Moreover,
the technique is simple to be implemented and the process is straightforward. 
Although it may increase the storage and inference time required in the training phase, these drawbacks are minor compared to the accuracy improvement expected.

Our proposed method used a CNN based object detection technique combined with image tiling technique, based on the leaf width of rice appeared in the image, 
to improve the performance of the detection model in case the object size in the target images vary. The leaf width in the image would be estimate by another dedicated CNN, trained by prior dataset.

\section{Materials and Methods}

This paper presented a pre-processing method based on CNN technique to minimize the effect of the object size variation in the image. 
The size of the leaf width in the image was estimated to determine the size of the sub images created by tile dividing. 
The image database with each leaf size calculated from polygon labeling were prepared as a training dataset for leaf width estimation model.
We used a CNN-based regression model as an architecture for the leaf width estimation model. We then used this model to predict the leaf width of all rice disease image database. 
The predicted leaf width value was used to decide the size that divided the original image from the database into sub-images via image tiling process.
The original images were then reconstructed into new image dataset that had similar size objects. This technique relied on automatic object size prediction to create a new set of training images similar to the augmentation technique that increased the number of images in model training. It improved the model performance by processing several object of similar size and enlarging small physical wounds more clearly. It also offered a flexible and convenient user experience because the user was not limited or controlled by the shooting distance or depended on the image resolution of the camera.

In this section, we presented materials preparation methods and our proposed technique details divided into three parts: 
1. Training Image Data Preparation,
2. Leaf Width Size Estimation Model Building, and 
3. Leaf Disease Detection Using Image Tiling Technique.

\subsection{Training Image Data Preparation}
\label{section:Data Acquisition}

\begin{figure}[hbtp!]
\begin{center}
\includegraphics[width=1.0\textwidth]{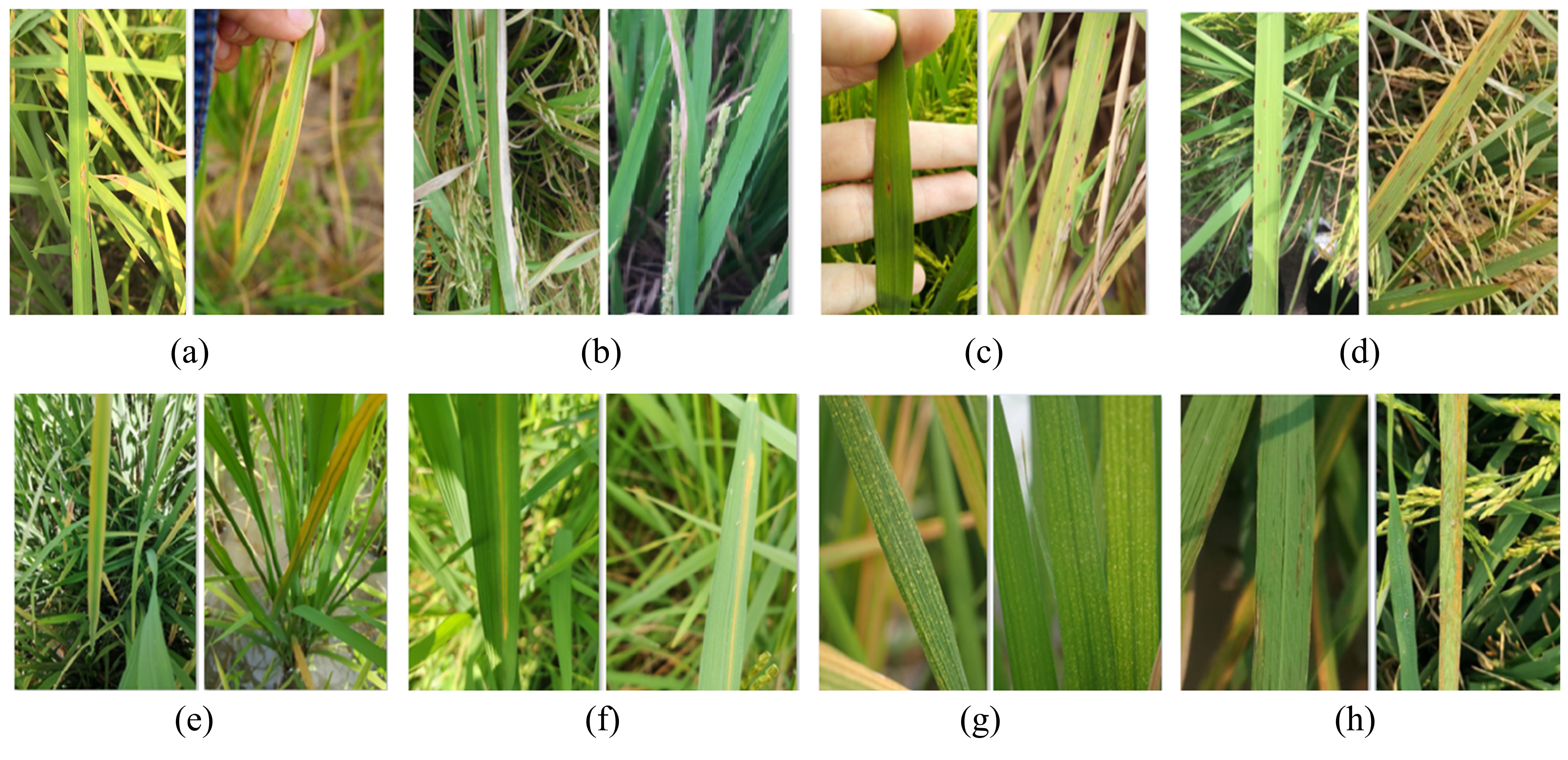}
\caption{Photographs of the eight rice diseases: (a) blast; (b) blight; (c) brown spot(BSP); (d) narrow brown spot(NBS); (e) orange leaf; (f) red stripe; (g) rice grassy stunt virus (RGSV); and (h) streak disease}
\label{fig:Fig_8disease.jpg}
\end{center}
\end{figure}

This subsection explains the preparation of rice disease images. 
We gathered a collection of the images of various disease types and stages after surveying the diseased rice fields in different areas in Thailand for several months. Figure.\ref{fig:Fig_8disease.jpg} shows the physical appearances of the eight main rice diseases found in Thailand including, (a) blast; (b) blight; (c) brown spot(BSP); (d) narrow brown spot(NBS); (e) orange leaf; (f) red stripe; (g) rice grassy stunt virus (RGSV); (h) streak disease. Each type of disease differs in color, size, shape, and texture. The diseases could be characterized by long thick stripes, short strokes, spots, and patches. The differences depended on the type and stages of the disease. The images were identified and screened by plant disease specialist who categorized by disease names. The whole leaf was labeled as an object area and the disease type.

\begin{figure}[hbtp!]
\begin{center}
\includegraphics[width=0.95\textwidth]{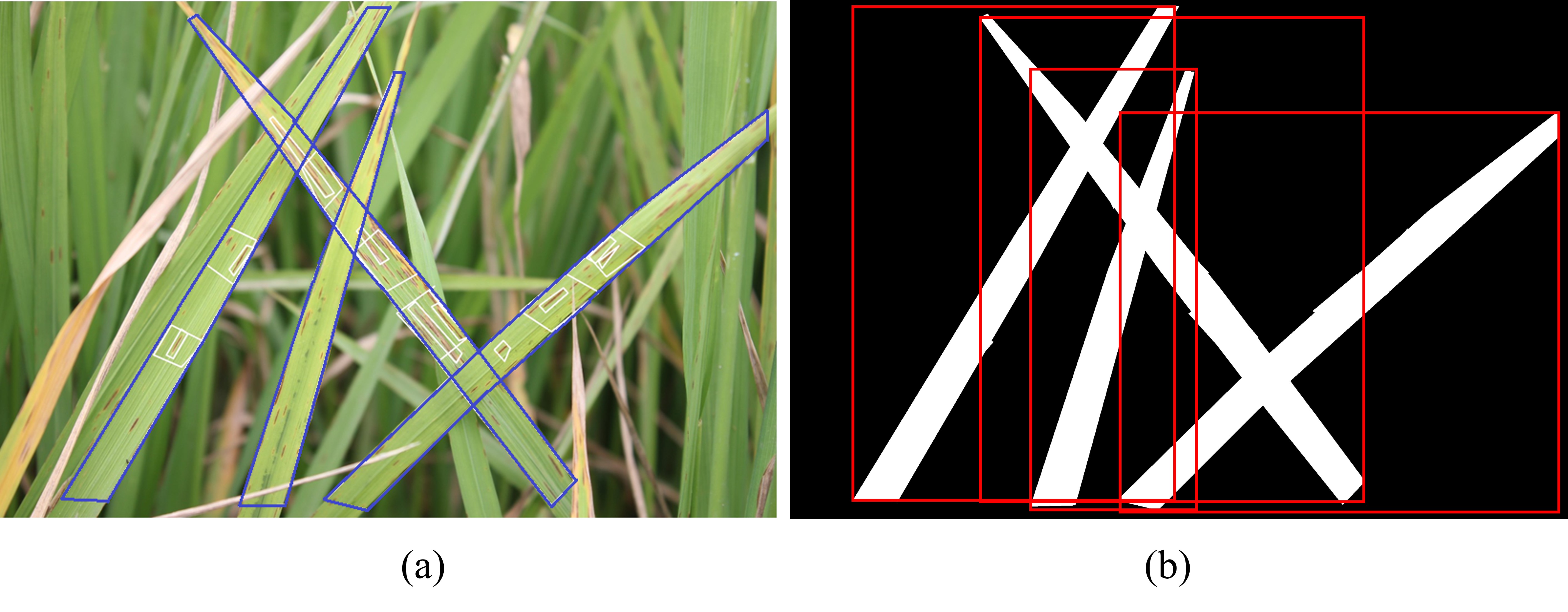}
\caption{Examples of polygons (blue line in (a)) and rectangle labeling image of rice leaves (red outlines in (b))}
\label{fig:Fig_polygon_mask.jpg}
\end{center}
\end{figure}

To obtain the leaf width in the image, polygon labeling annotation was applied.
As shown in Figure.\ref{fig:Fig_polygon_mask.jpg}, 
the rice leaf generally had long and narrow shape and the image had more than one leaf aligned in different direction.
We used the polygon labeling method instead of the regular rectangle labeling because polygon could provide leaf width information better than the rectangles, which could only roughly identify the area of interest.
Each object area was manually labeled by defining points around the object's region and the disease name appeared in the area. 
We calculated the leaf width information from this polygon labeling and trained a CNN model to estimate leaf width value in the images. The details are described in section \ref{section:method leaf width prepare}.

Furthermore, we used rectangle labeling for detection model as training image data because it took less time and less complicated CNN architecture than polygon labelling. Rectangular box presented enough information about object location in the image. Additionally, the position of the point from the polygon data could easily be converted to the rectangle coordinates.
We used LabelMe \cite{LabelMe} as an annotation tool for data labeling.

\subsection{Leaf Width Size Estimation Model Building}
This subsection described leaf width ground truth preparation and calculation method.
And how to use the prepared ground truth to train the leaf width size estimation model and its architecture.

\subsubsection{Leaf Width Ground Truth Preparation}
\label{section:method leaf width prepare}

In this section, the image processing techniques regarding connected component labeling, foreground/background segmentation and object shape fitting are described.
The ground truth of leaf disease regions prepared from section \ref{section:Data Acquisition} would be processed to calculate the size of the leaf width in the image. 

Firstly, the coordinates of the object region of the leaf area from the labeling information were used to separate foreground of each leaf object from the background.
Secondly, a fitting technique such as rectangle or elliptical shape was applied to fit each leaf foreground to determine the size of each object. In this process, binary image processing, connected component labeling techniques and the object coordinates data were applied to measure the dimensions of individual objects in the image.
Finally, the width size of the fitted object would be used to represent the leaf width size of the image. 
In case that there were multiple leaves in the image, the leaf width size was chosen from the largest leaf width. This value was normalized with the image's widest side in Eq.\ref{eq:leafval} to reduce the difference in a photo size factor.

\begin{equation} \label{eq:leafval}
\ LW = 100*\left ( \frac{lfw}{Max\left ( w,h \right )} \right )
\end{equation}
$LW$ is a normalized representative leaf width in the image in percentage, 
$lfw$ is leaf width size of image in pixels, and $w$ and $h$ are width and height of the image in pixels.

\subsubsection{Leaf Width Size Estimation Model Training}
\label{section:method leaf width train}
Although regular deep learning architectures need huge computational time and resources to train massive parameters, it depends on the size of the network and the number of the parameter to be trained.

Since leaf width size estimation was a simple regression task,
we focused on using small size CNN network.
We chose Resnet18 from one of the most popular ResNet (Residual Network) architectures \cite{ResNet_He_Zhang_Ren_Sun_2016} as a backbone network. Transfer learning, a well-known approach of deep learning method for saving time and resources, was chosen for model training.

Instead of starting the training from scratch, we initialized the network weights using a pre-trained network by ImageNet \cite{imagenet} dataset,
and adjusted the output layer from ImageNet's original object classification task to our regression object size estimation task.

\begin{figure}[hbt!]
\begin{center}
\includegraphics[width=0.9\textwidth]{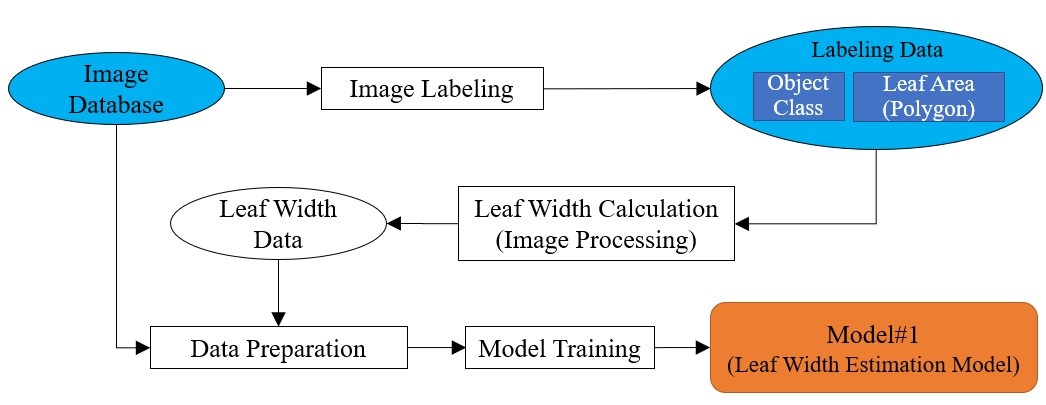}
\caption{A building process of leaf width estimation model }
\label{fig:flow process1.jpg}
\end{center}
\end{figure}

Image data prepared from section \ref{section:Data Acquisition} and each leaf width value calculated from section \ref{section:method leaf width prepare} were provided as ground truths data in order to assess the validity of the model performance. 
After completing the training with the desired accuracy performance, this model was used to automatically predict leaf size in other images. A flow process diagram of how to build a CNN model for leaf width estimation is shown in Figure.\ref{fig:flow process1.jpg}.
 
\subsection{Leaf Disease Detection Using Image Tiling Technique}
Tiled images preparation method and object detection model training and utilization methods used tiled images were introduced in this subsection subsequently.
 
\subsubsection{Tiled Images Preparation}
\label{section:method building tiled image}
The leaf width size of the image estimated by the model trained in section \ref{section:method leaf width train} was used to specify the size for image tile dividing.
In this process, the original input image would be cropped and divided into specified window size sub-images. 
The window was slid and split over the image from left-to-right and top-down directions. 
Set the window size to be bigger than twice the leaf size,
we defined an overlapping region between the sub-images, which we set to 50\% of the window size to make sure that the whole leaf in the horizontal direction would appeared in one or more sub-images.

\begin{figure}[hbt!]
\begin{center}
\includegraphics[width=0.95\textwidth]{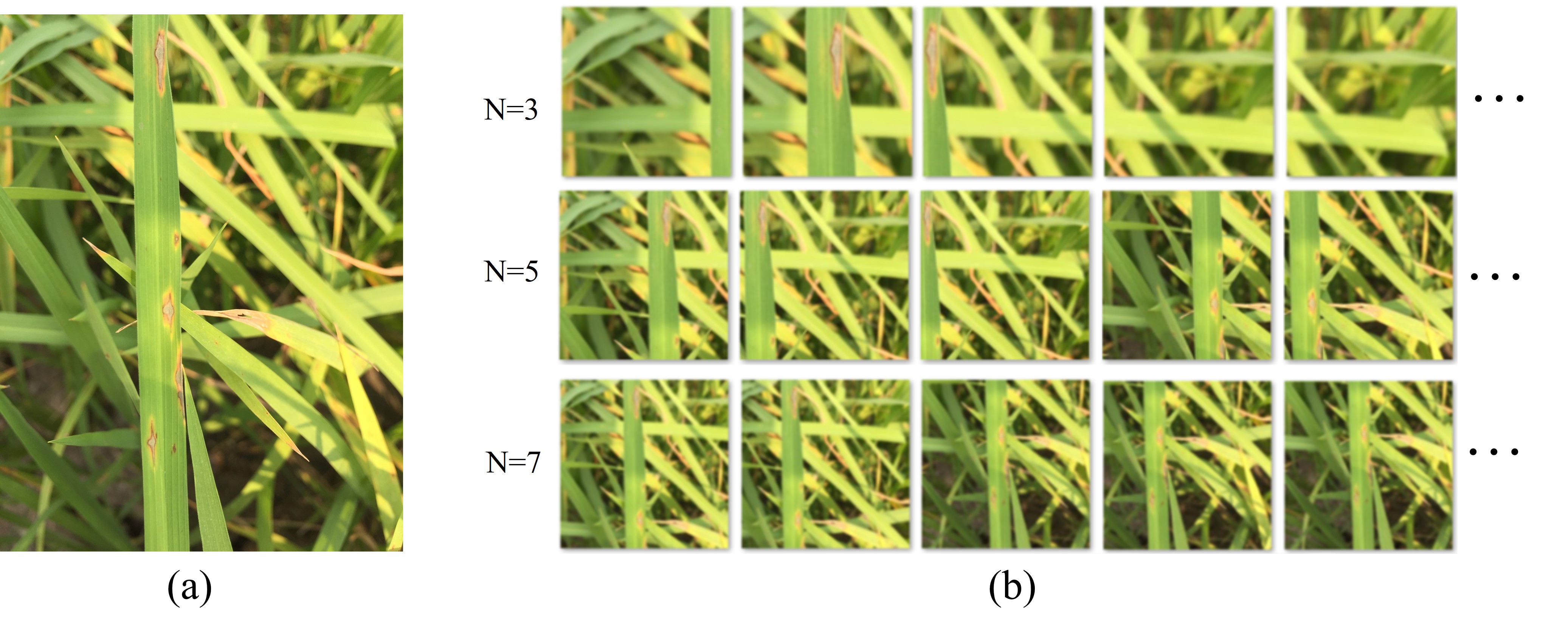}
\caption{Example of tiled sub-image: (a) input image, (b) result of tiled sub-image when N=3, 5, 7.}
\label{fig:Fig_Tiled_n357.jpg}
\end{center}
\end{figure}

The window size could be different depending on the characteristics of the object.
The size of the window was defined as N = 3, 5, 7 times the size of the leaf in our experiment, to find out which one provided the best model performance. 
In addition, if the aspect ratio of the object area in the sub-image to the ground truth area was less than the specified value, 
that sub-image would be discarded. 
The threshold criteria were determined to be less than 7\% empirically. 
Next, the ground truth coordinates on the original image were recalculated to re-position them relative to the coordinates on the new sub image. 
After the process was done, the obtained sub-images from the original input image would appear as tiled images as shown in the Figure.\ref{fig:Fig_Tiled_n357.jpg}. 
The Figure showed the result of tiled image processed from the input image in the left side of Figure.\ref{fig:Fig_8disease.jpg} (a). 

The new tiled images were used to trained the CNN rice disease prediction model.

\subsubsection{Object Detection Model Training and Utilization} 
\begin{figure}[hbt!]
\begin{center}
\includegraphics[width=1\textwidth]{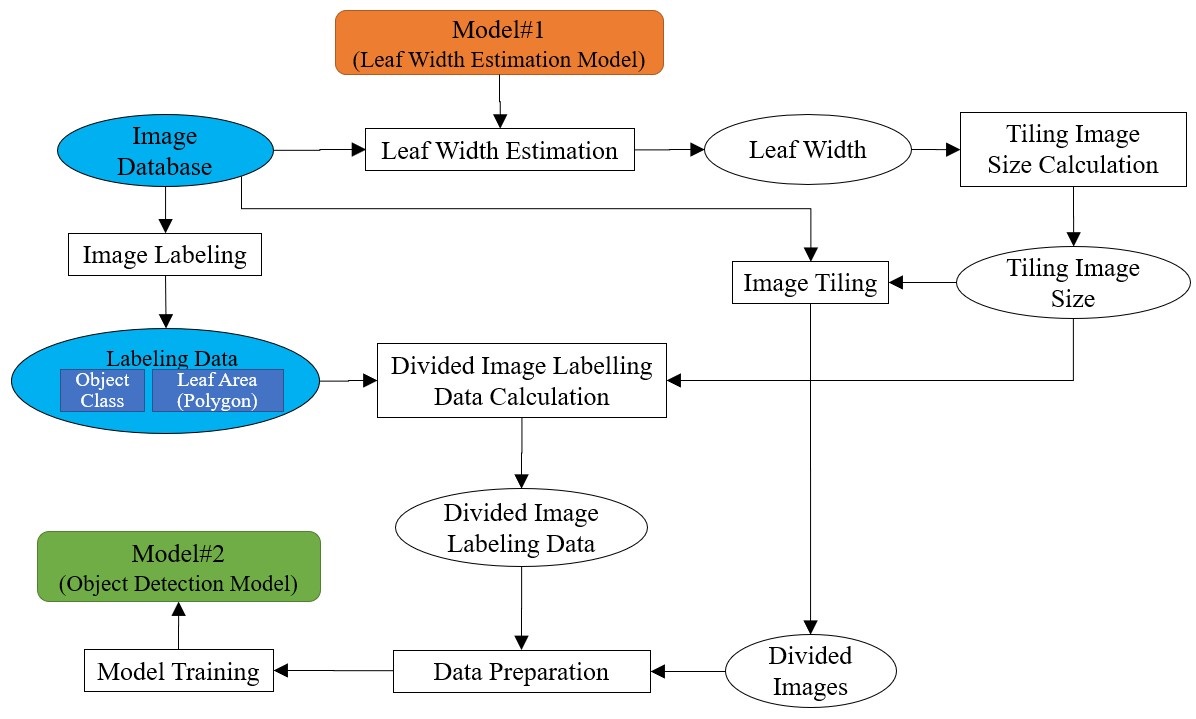}
\caption{A building process of object detection model for tiled image }
\label{fig:flow process2.jpg}
\end{center}
\end{figure}

\begin{figure}[hbt!]
\begin{center}
\includegraphics[width=0.8\textwidth]{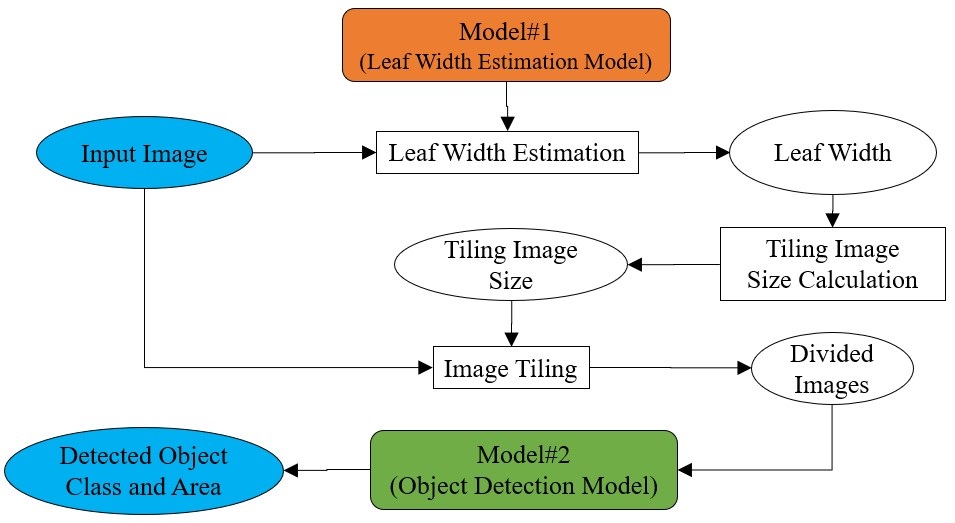}
\caption{Object detection utilization process using our proposed method}
\label{fig:flow process3.jpg}
\end{center}
\end{figure}

An object detection technique, YOLOv4 \cite{yolov4}, was adopted as a model to detect rice disease from tiled sub-image dataset. We chose YOLOv4 because its previous version YOLOv3 gave the best performance among various techniques in our previous study \cite{Kantip_2020}.
Flow process diagrams in Figure.\ref{fig:flow process2.jpg} and Figure.\ref{fig:flow process3.jpg} show how to build the modeling processes and how to use it to make predictions. The process consisted of several parts, which were described in detail in the previous sections.

We also made an experiment that trained YOLOv4 model using the same network parameter with the original input image to compare the performance and accuracy with our proposed method.

\section{Experiment Results and Discussions}
This section is
divided into three parts, 
1. Experimental Environment and Dataset,
2. Leaf Width Size Estimation Experiment Results, and
3. Performance of Leaf Disease Detection using image Tiling Technique.
In this experiment, single shot CNN object detection technique, YOLOv4 \cite{yolov4}, was used as baseline model to detect rice leaf disease.

\subsection{Experimental Environment and Dataset}

The developed technique used C++ programming and Python language, and the experiment was run under Linux operating system on 
NVIDIA DGX1, Dual 20-Core Intel Xeon E5-2698 v4 2.2 GHz CPU, 512 GB 2,133 MHz, 8X NVIDIA Tesla V100 GPU.
In our dataset, there were up to 20,008 images of eight rice leaf diseases.
The proportion of the number of images in each class was varied depending on the survey and the frequency of the disease, as shown in Table.\ref{table:8class_dataset}.

\begin{table}[hbt!]
\begin{tabular}{|l|c|c|} 
\cline{1-3}
\textbf{Class Name} & \multicolumn{1}{l|}{\textbf{No. of Image Data}} & \multicolumn{1}{l|}{\begin{tabular}[c]{@{}l@{}}\textbf{Image Data}\\\textbf{Percentage (\%)}\end{tabular}}  \\ 
\cline{1-3}
Blast               & 3,041                                           & 15.20                                                                                                         \\ 
Blight              & 3,466                                           & 17.32                                                                                                         \\ 
BSP                 & 1,735                                           & 8.67                                                                                                          \\ 
NBS                 & 2,356                                           & 11.78                                                                                                         \\ 
Orange              & 921                                             & 4.60                                                                                                          \\ 
Red                 & 2,575                                           & 12.87                                                                                                         \\ 
RGSV                & 3,981                                           & 19.90                                                                                                         \\ 
Streak              & 1,933                                           & 9.66                                                                                                          \\ 
\cline{1-3}
\textbf{Total}      & \textbf{20,008}                                 & \textbf{100}                                                                                                  \\
\cline{1-3}
\end{tabular}
\centering
\caption{Datasets of eight rice disease in our study}
\label{table:8class_dataset}
\end{table}

\subsection{Leaf Width Size Estimation Experiment Results}
This subsection showed experiment results involved with leaf width size estimation model.
It was divided into two parts, Leaf Width Size Estimation Model's Performance, and Leaf Disease Detection Performance Analysis Based on Leaf Width Size.

\subsubsection{Leaf Width Size Estimation Model's Performance}
\label{section:expI}
The experiment was conducted to train and measure the effectiveness of the CNN model used to estimate the rice leaf width in the image. To save computational resources and avoid effect from disease class imbalance problem, the images in the dataset were randomly chosen and sampled equally at 620 images per class. A total of 4,960 images of eight types of rice leaf disease and the images was studied. The image were divided into 3 sets (training, validation and test sets) in the proportion of 80:10:10, respectively. 
The ground truth of the leaf width value was calculated using the method described in section \ref{section:method leaf width prepare}, in order to be used in model training process. 
The ResNet18 architecture, a small 18-layer network structure, was chosen 
to solve the regression leaf width estimation problem.

The model returned a percentage of the relative leaf width with respect to the image longest side length. The simple Mean Absolute Error (MAE) was chosen as a loss function and the model that performed the best performance during 100 epochs training was chosen to be used in the later process.

\begin{figure}[hbt!]
\begin{center}
\includegraphics[width=0.75\textwidth]{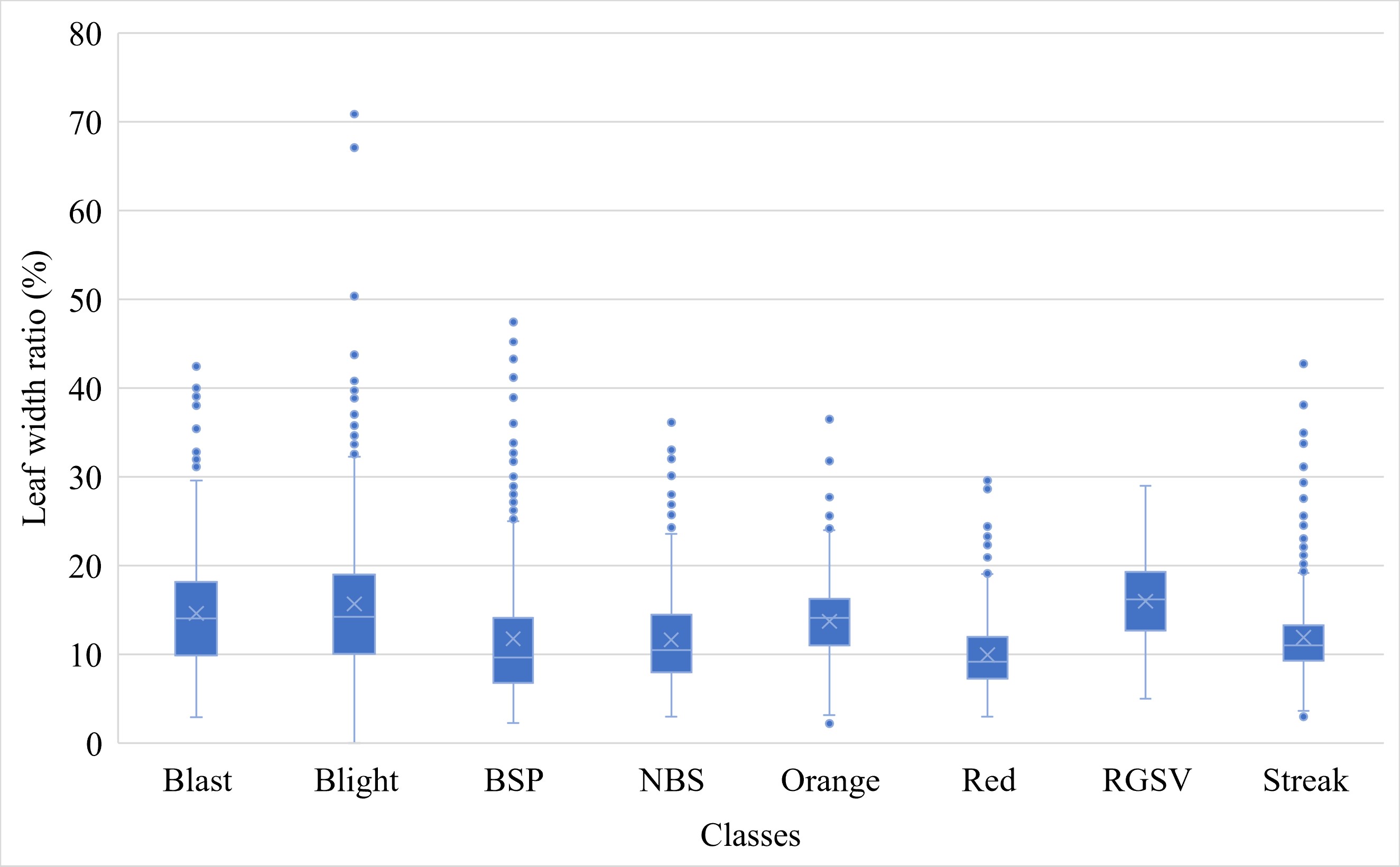}
\caption{Box plot of leaf width ratio ground truth value on the eight rice leaf diseases entire dataset}
\label{fig:graph leaf width of ground truth.jpg}
\end{center}
\end{figure}

\begin{figure}[hbt!]
\begin{center}
\includegraphics[width=0.8\textwidth]{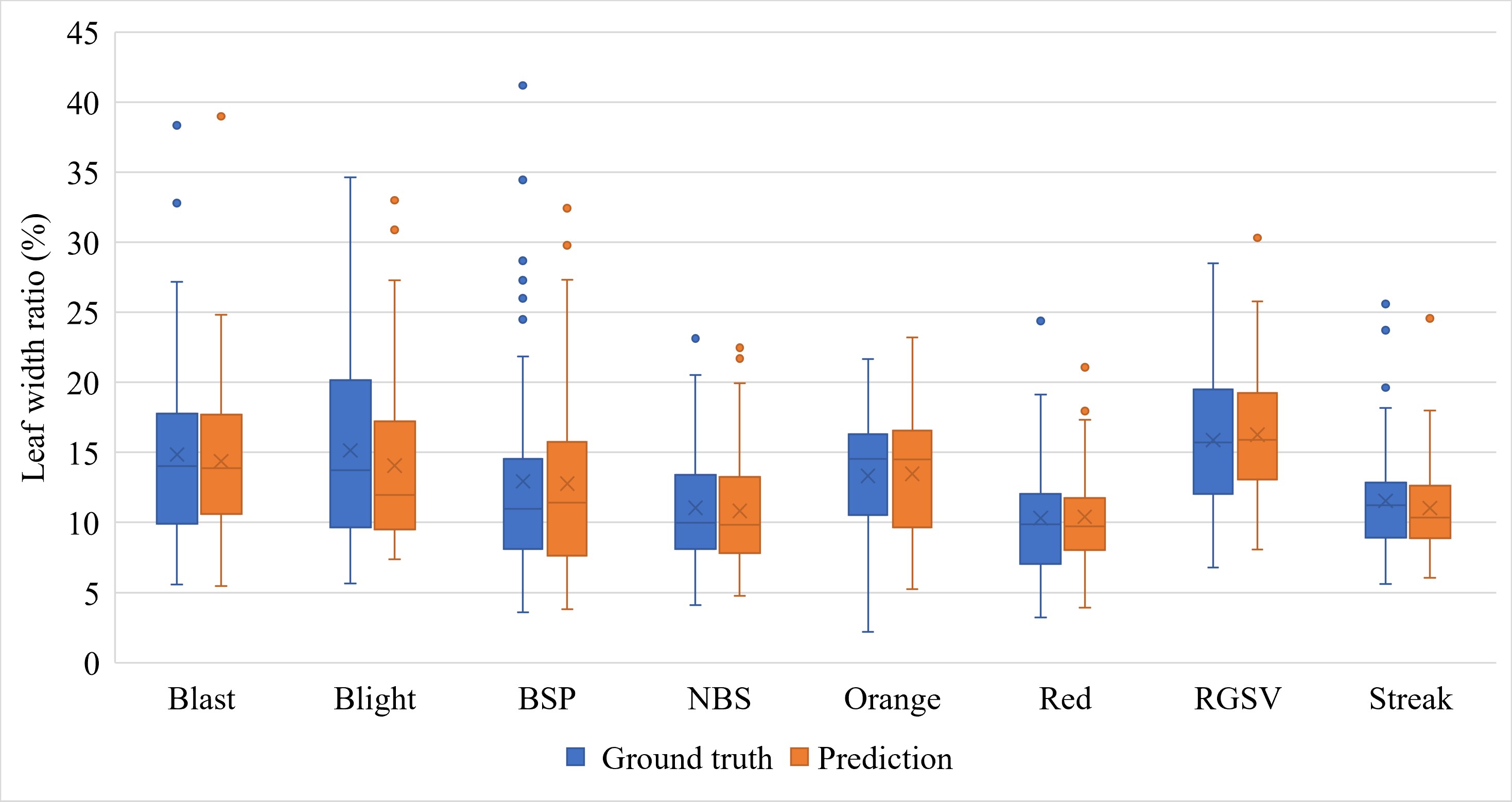}
\caption{Box plot comparison between ground truth and prediction results of leaf width ratio value on the eight rice leaf diseases test dataset}
\label{fig:Fig boxplot compare result.jpg}
\end{center}
\end{figure}

We evaluated the model on test set of 496 images.
The leaf width value was visualized using a boxplot with statistics including Max, Min, Mean and SD to describe its characteristics. The distributions of ground truth data of all images in our dataset are shown in Figure.\ref{fig:graph leaf width of ground truth.jpg} while the performance comparison on the test dataset between a predicted result and the actual ground truth is showed in Figure.\ref{fig:Fig boxplot compare result.jpg}. To analyze the performance, we used a mean absolute percentage error (MAPE) in Eq.\ref{eq:MAPE}. to measure the accuracy of prediction results.

The error rates of each disease class were 10.25\%, 11.35\%, 14.99\%, 10.37\%, 12.45\%, 12.42\%, 10.28\%, 7.32\% for blast, blight, brown spot, narrow brown spot, orange, red strip, RGSV, streak class, respectively. The average error of all 8 classes of the leaf width prediction model was 11.18\%, which was sufficiently accurate for the usage of tiling sub-image size determination.  

\begin{equation} \label{eq:MAPE}
 MAPE=\frac{1}{n} \sum_{i=1}^{n}\left | \frac{_{A_{t}-F_{t}}}{A_{t}} \right |
\end{equation}

where $F_{t}$ is the ground truth value, $A_{t}$ is the predicted value and $n$ is the number of images.

\subsubsection{Leaf Disease Detection Performance Analysis Based on Leaf Width Size}
\label{section:expII}
\begin{figure}[hbt!]
\begin{center}
\includegraphics[width=0.8\textwidth]{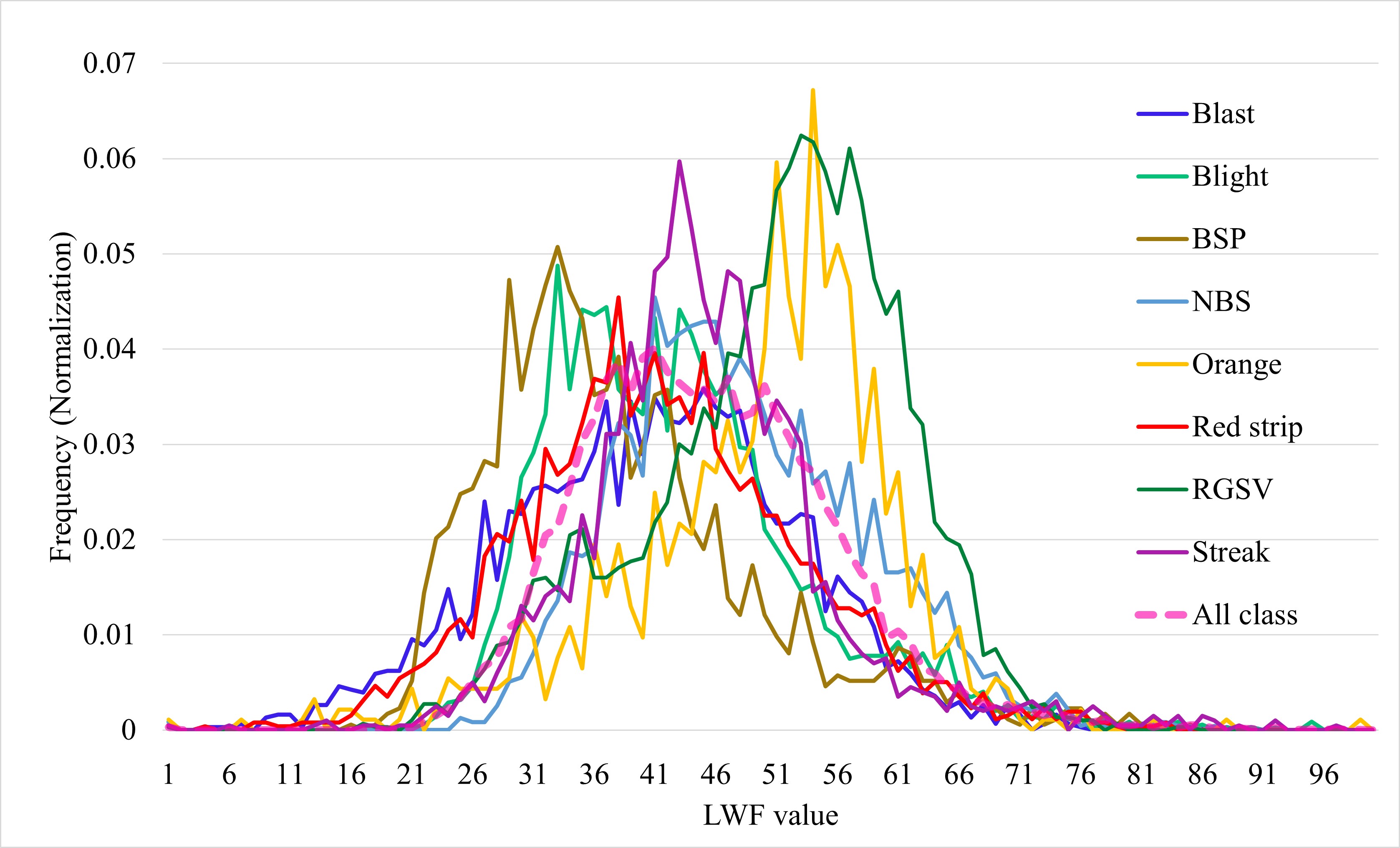}
\caption{The leaf width characteristics in the images in our dataset obtained from the prediction model and measured by the T-score distribution.}
\label{fig:Fig Tscore Bell Curve}
\end{center}
\end{figure}

To study the performance of the leaf disease prediction model affected by variations in object size, all images in our dataset were examined. 
The leaf width prediction model from section \ref{section:expI} was used to estimate leaf width in the image in order to differentiate between images of normal sized leaves and images of deviated uncommon sized leaves.
For shooting wide-angle/narrow-angle subjects, it could be narrow or wide relative to the body of the subject. In this work, the narrow or wide rice leaf referred to the proportion of the leaf size compared to the image size in the photograph.
To identify leaves that are narrow or wide apart from their normal size, the T-score method was used to represent distances between sample means and the population mean. Figure.\ref{fig:Fig Tscore Bell Curve} shows a bell-shaped histogram obtained by plotting a frequency graph of occurrence of leaf width values. 

This value was set to $LWF$ calculated from the value referenced from T-score as shown in Eq.\ref{eq:Tscore}, with the total area under the distribution curve equal to 1. Let, the T-score of the sample with value $x$ be $T(x)=\sqrt{x - \mu}$ and $T_{min} = min_{i=1}^{n}T(x(i))$, when $n$ = total number of each class sample, $x(i)$ = $LW$ value of the sample $i$, and $\mu$ = estimated mean of $LW$ value. 

\begin{equation} \label{eq:Tscore}
LWF = \sqrt{T(x)-T_{min}}
\end{equation}

\begin{figure}[hbt!]
\begin{center}
\includegraphics[width=0.8\textwidth]{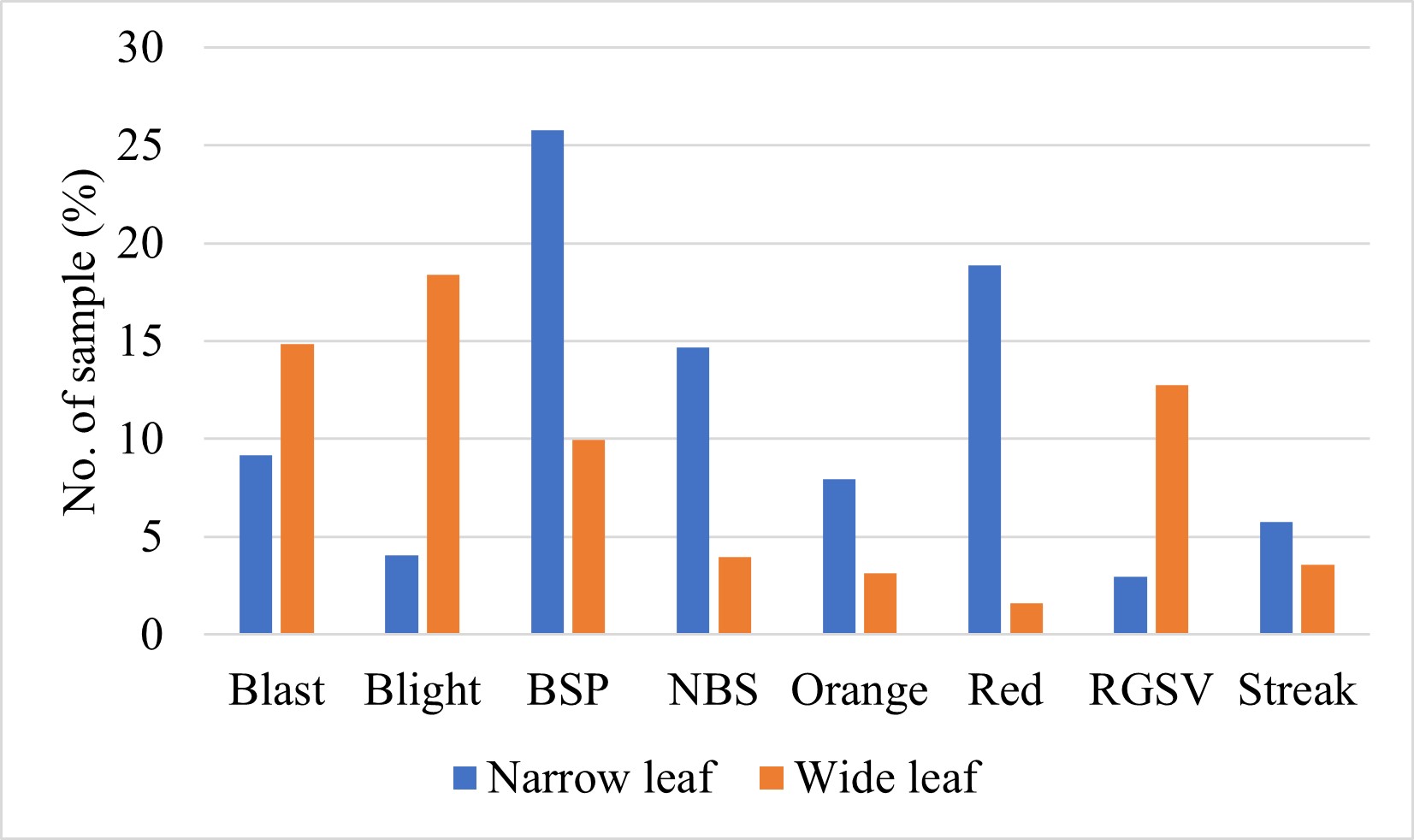}
\caption{The proportion of the number of images in each class divided by leaf width.}
\label{fig:graph t score data proportion.jpg}
\end{center}
\end{figure}

Most of the samples were close to the mean value. 
The lower and upper tails on the left and right sides of the curve were the narrow and the wide leaf width values. To separate deviated sized leaves from normal sized leaves, we calculated the proportion of the number of samples from the bell curve distribution to divide them into three groups including a sample of 10\% narrow-sized leaf on the left tail, a sample of 10\% wide-sized leaf on the right tail, and the other 80\% normal-sized leaf from the middle bell curve. The number of image samples obtained from each class had different proportions as shown in Figure.\ref{fig:graph t score data proportion.jpg}. The classified data was then used to train the model and evaluate the prediction results. The dataset of normal-sized leaves was used to build a model for predicting disease types to regenerate the situation of having inadequate leaf size variety in the training dataset and use as a baseline. This data set was further divided into three sub-sets (training, validation and test set) with proportions of 80\%, 10\%, 10\%, respectively. 

\begin{table}[hbt!]
\centering
\begin{tabular}{|l|c|c|c|} 
\cline{2-4}
\multicolumn{1}{l|}{}                          & \multicolumn{3}{c|}{\textbf{Average Precision (AP)}}                                                                               \\ 
\cline{1-4}
\textbf{\textbf{Class~\textbf{\textbf{Name}}}} & \multicolumn{1}{l|}{\textbf{Narrow leaf}} & \multicolumn{1}{l|}{\textbf{Normal leaf}} & \multicolumn{1}{l|}{\textbf{Wide leaf}}    \\ 
\cline{1-4}
Blast                                          & 80.38                                     & 93.04                                     & 95.77                                      \\ 
Blight                                         & 75.92                                     & 98.58                                     & 94.07                                      \\ 
BSP                                            & 51.26                                     & 87.96                                     & 97.87                                      \\ 
NBS                                            & 63.5                                      & 96.26                                     & 99.91                                      \\ 
Orange                                         & 14.58                                     & 86.91                                     & 96.31                                      \\ 
Red                                            & 95.76                                     & 99.77                                     & 93.67                                      \\ 
RGSV                                           & 70.87                                     & 96.37                                     & 98.88                                      \\ 
Streak                                         & 80.2                                      & 94.38                                     & 94.67                                      \\ 
\cline{1-4}
\textbf{mAP}                                   & \textbf{66.56}                            & \textbf{94.16}                            & \textbf{96.39}                             \\
\cline{1-4}
\end{tabular}

\caption{Average precision (AP) of the model tested in three groups of images categorized by leaf width size}
\label{table:test3group}
\end{table}

\begin{figure}[hbt!]
\begin{center}
\includegraphics[width=0.8\textwidth]{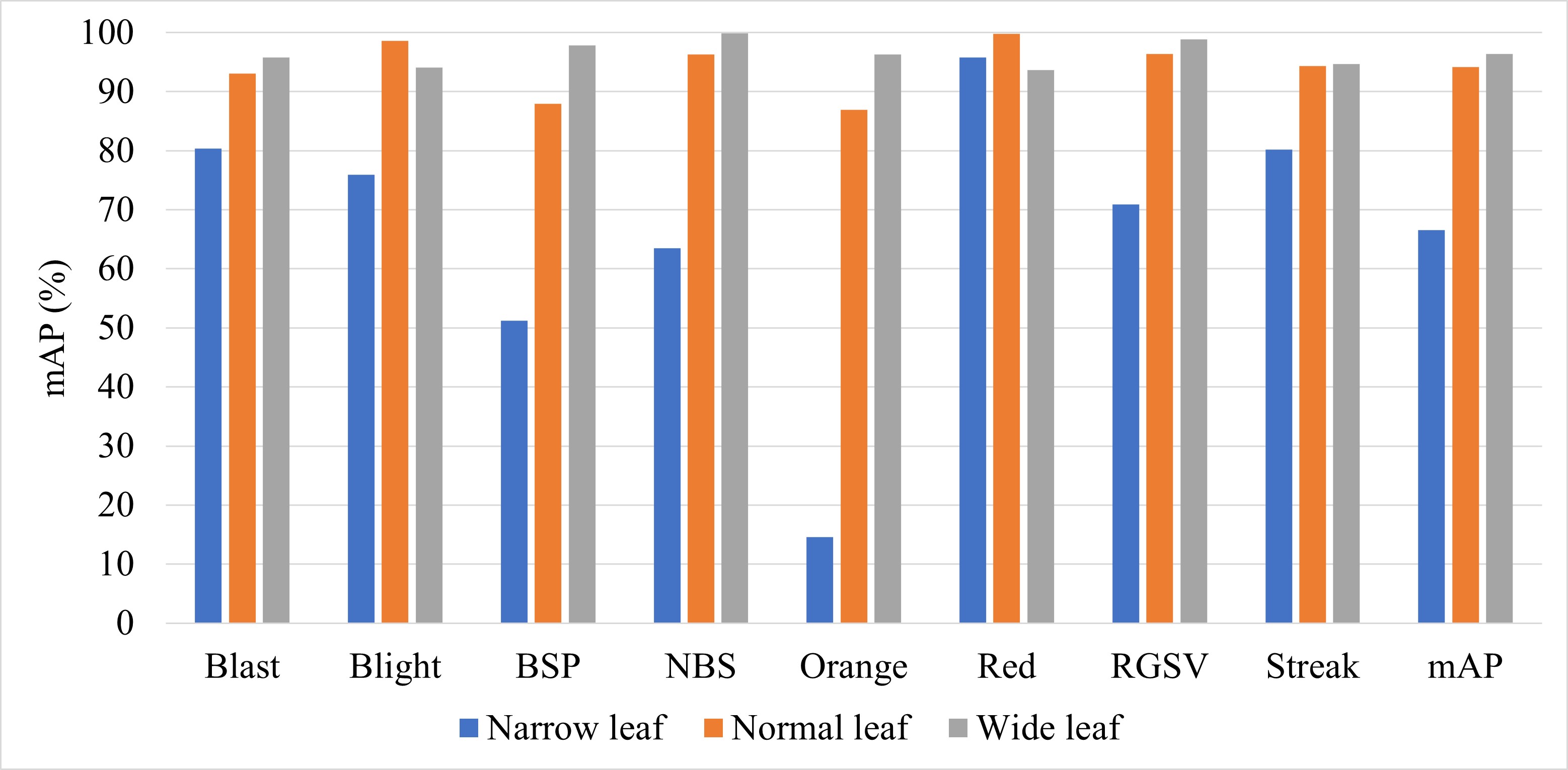}
\caption{Average precision (AP) of the baseline object detection model tested in three groups of images categorized by leaf width size}
\label{fig:graph t score mAP.jpg}
\end{center}
\end{figure}
Table.\ref{table:test3group} and Figure.\ref{fig:graph t score mAP.jpg} show the test dataset predicted result on normal-sized leaf was at 94.16\% while the wide-sized leaf was 96.39\%, higher than the normal-sized leaf. The result on narrow-sized leaf was significantly lower, at only 66.56\%. It was clear that different object sizes influenced the predictive performance. This was especially noticeable in the case of small objects in the image, which was 27.6\% lower than the baseline.

\subsection{Performance of Leaf Disease Detection Using Image Tiling Technique}
\label{section:expIII}

The data in this experiment were prepared as described in section \ref{section:expI}. The total number of 4,960 images was divided into training, validation and test sets in the proportion of 80:10:10, respectively. 
The photographs from the original image dataset were converted to a tiled sub-images dataset based on the leaf width in the image by using the processes described in section 
\ref{section:method building tiled image}. 
In the tiled images dataset, the object in the image was processed to adjust their proportions relative to the calculated leaf width. To study the effect of difference in object size in the image, the two datasets from original images and their tiled images were then used to trained YOLOv4 models with the same configuration to compare their performance. 

A comparative study was conducted to determine the best tiled sub-image size. Let $N$ be the coefficient number of the leaf width value varied between 3, 5, 7, the sub-image size was determined as $N \times LW$.
In Table.\ref{table:tilede data set}, which shows the image dataset of the original input image and tiled set images at $N$ values, the number of images in the tiled set was increased by 1.5-4.4 times more than the original input image set. There were 21864, 12063, and 7467 images for $N$ = 3, 5, 7 respectively, for a total of 41,394 images.

\begin{comment}
\begin{table}[hbt!]
\begin{center}
\includegraphics[width=0.65\textwidth]{fig/Fig_Data_set_numImage.jpg}
\caption{Dataset of tiled image of eight rice diseases.}
\label{table:Fig_Data_set_numImage.jpg}
\end{center}
\end{table}
\end{comment}

\begin{table}
\centering
\begin{tabular}{|l|c|c|c|c|} 
\hline

\textbf{Class} & \textbf{No. of } 
&\multicolumn{3}{c|}{\textbf{No. of Tiled Image }}   \\ 

\cline{3-5} \textbf{Name} & \textbf{Original Image} & \textbf{N=3}    & \textbf{N=5}    & \textbf{N=7}    \\ 
\hline
Blast                                                                                   & 620                                                                                                & 2,533           & 1,379           & 955             \\ 
Blight                                                                                  & 620                                                                                                & 2,631           & 1,532           & 871             \\ 
BSP                                                                               & 620                                                                                                & 2,656           & 1,380           & 840             \\ 
NBS                                                                                     & 620                                                                                                & 2,552           & 1,373           & 860             \\ 
Orange                                                                                  & 620                                                                                                & 2,987           & 1,533           & 1,131           \\ 
Red                                                                                     & 620                                                                                                & 2,862           & 1,667           & 896             \\ 
RGSV                                                                                    & 620                                                                                                & 2,950           & 1,550           & 1,219           \\ 
Streak                                                                                  & 620                                                                                                & 2,693           & 1,649           & 695             \\ 
\hline
\textbf{Total}                                                                          & \textbf{4,960}                                                                                     & \textbf{21,864} & \textbf{12,063} & \textbf{7,467}  \\
\hline
\end{tabular}
\caption{Dataset of tiled image of eight rice diseases.}
\label{table:tilede data set}
\end{table}

\begin{figure}[hbt!]
\begin{center}
\includegraphics[width=0.9\textwidth]{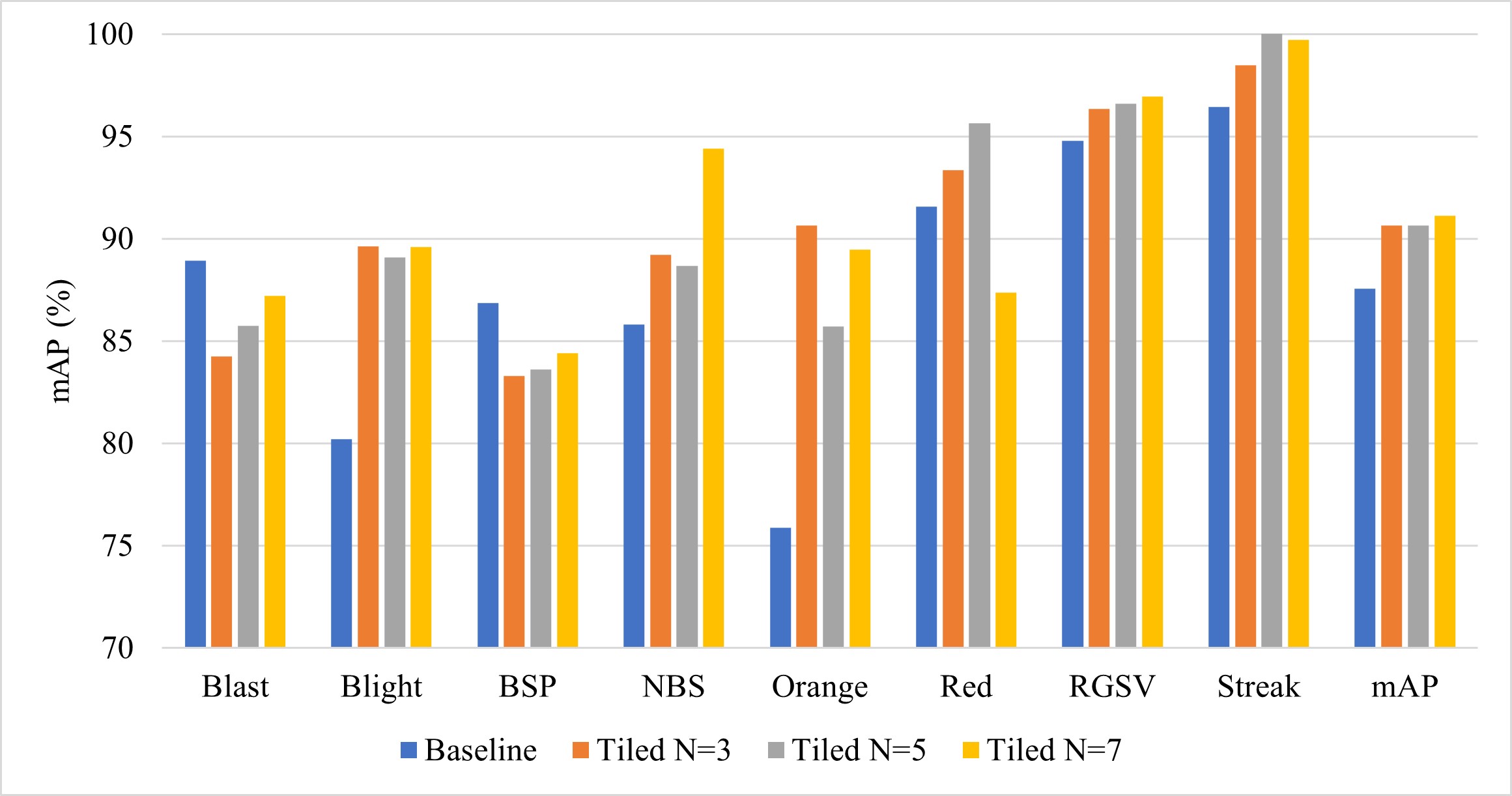}
\caption{Average precision (AP) of the detecting model on test data set with/without tiling for detecting eight rice diseases.}
\label{fig:Fig_mAP8class_train_test_bar_chart.jpg}
\end{center}
\end{figure}

In Table.\ref{table:mAP on tiled data set} and Figure.\ref{fig:Fig_mAP8class_train_test_bar_chart.jpg}, the performance of the model trained on the tiled image set had higher performance on mAP over all N values compared to the model trained on the original image set. The best result on the tiled set was 91.14\% ($N$=7), which was higher than 87.56\% obtained on the original image set. In each class, such as blight, orange and narrow brown spot disease, there was a significant improvement. The efficiency on blight and orange leaf disease increased by more than 9.4\% and 14.8\%, respectively, when $N$=3, while NBS increased by more than 8.6\% when N=7. However, there were two diseases types that had lower precision, i.e. blast and brown spot disease, where the AP decreased by 1.7\%-4.7\% and 2.5\%-3.6\%, respectively, for each $N$ value. 

\begin{comment}
\begin{table}[hbt!]
\begin{center}
\includegraphics[width=0.90\textwidth]{fig/Fig_mAP8class_train_test_table.jpg}
\caption{Average precision (AP) of the detecting model with/without tiling for detecting eight rice diseases.}
\label{table:Fig_mAP8class_train_test_table.jpg}
\end{center}
\end{table}
\end{comment}

\begin{table}[hbt!]
\centering
\begin{tabular}{|l|c|c|c|c|c|c|c|c|} 
\hline

\textbf{Class} & \multicolumn{2}{c|}{\textbf{Original}} & \multicolumn{6}{c|}{\textbf{Tiled Image Set}} \\
\cline{4-9} \textbf{Name} & \multicolumn{2}{c|}{\textbf{Image Set}} & \multicolumn{2}{c|}{\textbf{N=3}} & \multicolumn{2}{c|}{\textbf{N=5}} & \multicolumn{2}{c|}{\textbf{N=7}} \\

\cline{2-9}
                                                                                        & \textbf{train} & \textbf{test}                                                                                        & \textbf{train} & \textbf{test}    & \textbf{train} & \textbf{test}    & \textbf{train} & \textbf{test}     \\ 
\hline
Blast                                                                                   & 94.86          & 88.94                                                                                                & 91.64          & 84.25            & 93.84          & 85.75            & 94.77          & 87.22             \\ 
Blight                                                                                  & 93.88          & 80.21                                                                                                & 94.19          & 89.63            & 95.76          & 89.1             & 95.87          & 89.6              \\ 
BSP                                                                                     & 74.97          & 86.86                                                                                                & 83.35          & 83.29            & 82.86          & 83.61            & 74.42          & 84.4              \\ 
NBS                                                                                     & 87.44          & 85.82                                                                                                & 90.4           & 89.21            & 94.07          & 88.67            & 88.84          & 94.39             \\ 
Orange                                                                                  & 77.53          & 75.87                                                                                                & 94.73          & 90.64            & 89.63          & 85.71            & 79.65          & 89.46             \\ 
Red                                                                                     & 97.87          & 91.56                                                                                                & 92.46          & 93.34            & 93.25          & 95.66            & 96.18          & 87.37             \\ 
RGSV                                                                                    & 93.57          & 94.79                                                                                                & 96.2           & 96.36            & 94.73          & 96.6             & 95.36          & 96.95             \\ 
Streak                                                                                  & 96.77          & 96.45                                                                                                & 96.91          & 98.47            & 97.14          & 100              & 99.04          & 99.71             \\ 
\hline
\textbf{mAP}                                                                            & \textbf{89.61} & \textbf{87.56}                                                                                       & \textbf{92.48} & \textbf{90.65}   & \textbf{92.66} & \textbf{90.64}   & \textbf{90.52} & \textbf{91.14}    \\
\hline
\end{tabular}
\caption{Average precision (AP) of the detecting model with/without tiling for detecting eight rice diseases.}
\label{table:mAP on tiled data set}
\end{table}

For $N$ value, it was unclear which $N$ gave the best performance because $N$ yielded quite different results in some diseases. NBS had a significantly higher AP when $N$ was 7, as opposed to red leaf disease, in which the AP was lower. In the case of orange leaf disease where $N$ = 5, the result was abnormally low in comparison to the other $N$ values. 

In term of the physical characteristics of the leaf disease image, the size of the wound area or the stage of the disease that appeared on the leaves could affect the model effectiveness. On the disease where wound areas were visible throughout the leaves, such as blight, red, orange, RGSV, streak disease, the model performed better than on the disease with wound area appeared only partially across the leaf, such as blast and brown spot disease. 
The reason why these disease groups outperformed others was suspected to be because there was a higher chance of wound appearing in the sub-image compared to the disease with the wound appeared all over the leaves.

We have shown that the developed technique improved the efficiency of rice disease detection. This technique used automatic leaf width prediction to adjust the size of the object in the image relative to the width of the leaf in the photograph. It reduced the effects of the leaf size difference from the photos and it made the small wounds became more visible. In addition, similar to the augmentation technique, it increased the amount of image data used to train the model to learn more diverse data. It also allowed for more flexible and convenience for users to photograph due to fewer restrictions on shooting distance or image resolution.

From this study, we suggested the following points to further improve the technique. 
In the future, this technique may be adapted to a certain range of object sizes instead of applying it to all images. Another issue worth investigating is an ability to detect small or large objects in the same frame or in a certain area.
In addition, it should be extended to cover many other rice diseases where symptoms may appear in other parts of the plant to further improve the efficiency of our Line Bot diagnostic system. Currently, the system provides services for a while and has increased users from rice planting areas in many regions in several provinces of Thailand.

\section{Conclusions}
This paper presented a tiling image division using an automatic width leaf estimation technique to deal with leaf size differences in rice disease photographs taken in field conditions. Our experiment was evaluated on a database of eight types of rice leaf diseases. We found that the technique for leaf width estimation model based on the ResNet18 architecture achieved a high performance estimated with MAPE at 11.18\% on 4,960 images. 
The leaf width model was used to categorize over 20,000 images according to leaf width size into three subgroups in order to assess the performance of leaf disease predictions on the effects of various object sizes. The performance of YOLOv4 model showed that mAP detection rate in small-sized leaves was significantly reduced to only 66.56\% compared to 94.16\% of normal-sized leaves, while wide-sized leaves increased detection rates up to 96.39\%. 
To reduce the problem of this effect, the tiling image division technique was applied to improve the objects in the image to be more similar in size. 
This experiment was evaluated on a tiled image set of 1.5-4.4 times the original input image set of 4,960 images.
The presented technique provided the best mAP detection performance at 91.14\% compared to the baseline, which was 87.56\%. It could improve 6 of the 8 diseases. 
Further investigations would improve the technique performance and expand to cover other rice diseases.

\section{Acknowledgement}
This study was a part of Mobile application for rice disease diagnosis using image analysis and artificial intelligence" project grant no. P18-51456, supported by grants from Innovation for Sustainable Agriculture (ISA), National Science and Technology Development Agency (NSTDA), Thailand. We would like to thank the team from Department of Plant pathology, Faculty of Agriculture at Kampheng Sean, Kasetsart University, who supported our data preparation and provided plant disease diagnosis knowledge. The authors are thankful to Prof. S. Seraphin, NSTDA Professional Authorship Center, for fruitful discussion on manuscript preparation and English editing.

\end{document}